\documentclass[conference]{IEEEtran}
\IEEEoverridecommandlockouts
\usepackage{cite}
\usepackage{amsmath,amssymb,amsfonts}
\usepackage{algorithmic}
\usepackage{graphicx}
\usepackage{textcomp}
\usepackage{xcolor}
\usepackage{hyperref}
\def\BibTeX{{\rm B\kern-.05em{\sc i\kern-.025em b}\kern-.08em
    T\kern-.1667em\lower.7ex\hbox{E}\kern-.125emX}}
\begin{document}

\title{DiffListener: Discrete Diffusion Model \\ for Listener Generation\thanks{This work was supported by Institute of Information \& communications Technology Planning \& Evaluation (IITP) grant funded by the Korea government (MSIT) (No.2022-0-00608, Artificial intelligence research about multi-modal interactions for empathetic conversations with humans \& No.RS-2020-II201336, Artificial Intelligence graduate school support(UNIST)) and the National Research Foundation of Korea(NRF) grant funded by the Korea government(MSIT) (No. RS-2023-00219959).}}

\author{\IEEEauthorblockN{Siyeol Jung, Taehwan Kim}
\IEEEauthorblockA{Artificial Intelligence Graduate School, UNIST, Republic of Korea \\
\{siyeol, taehwankim\}@unist.ac.kr}
}

\maketitle
\begin{abstract}
The listener head generation (LHG) task aims to generate natural nonverbal listener responses based on the speaker's multimodal cues.
While prior work either rely on limited modalities (e.g. audio and facial information) or employ autoregressive approaches which have limitations such as accumulating prediction errors.
To address these limitations, we propose DiffListener, a discrete diffusion based approach for non-autoregressive listener head generation. 
Our model takes the speaker's facial information, audio, and text as inputs, additionally incorporating facial differential information to represent the temporal dynamics of expressions and movements. 
With this explicit modeling of facial dynamics, DiffListener can generate coherent reaction sequences in a non-autoregressive manner.
Through comprehensive experiments, DiffListener demonstrates state-of-the-art performance in both quantitative and qualitative evaluations. 
The user study shows that DiffListener generates natural context-aware listener reactions that are well synchronized with the speaker.
The code and demo videos are available in \url{https://siyeoljung.github.io/DiffListener}.
\end{abstract}
\begin{IEEEkeywords}
Listener Head Generation, Discrete Diffusion, Dyadic Conversation
\end{IEEEkeywords}

\section{Introduction}
With the rise of applications such as digital avatar generation~\cite{chen2021high,li2021ai,zhang2022motiondiffuse} and human-computer interaction~\cite{zhang2021flow,guo2021ad,peng2023emotalk}, 
appropriate listening reactions have attracted attentions~\cite{tolzin2023mechanisms,cho2022alexa}.
In face-to-face conversations, appropriate nonverbal feedback from the listener, rather than content-based replies, is often crucial for maintaining the flow of communication~\cite{cassell1999power}. 
Therefore appropriate nonverbal feedback is important for virtual agents that communicate with humans~\cite{tronick1980monadic}.
In this context, there is growing interest in generating realistic listener head generation (LHG)~\cite{ng2022learning,ng2023can,song2023emotional}.
The listener head generation task aims to generate natural and appropriate nonverbal responses, such as head motions and facial expressions, based on the speaker's verbal utterances and facial expressions.
Moreover, listener reactions are influenced not only by the speaker's input but also by the listener's internal state, emotions, and personality, leading to non-deterministic responses.
To address this non-deterministic property of the listener's reaction, a previous study~\cite{ng2022learning} proposes to model the listener's reactions using a one-dimensional VQ-VAE~\cite{van2017neural}. 
This allows modeling unique reactions for different listener identities while preserving the non-deterministic properties and identity-specific reaction styles.

Most existing LHG approaches rely on autoregressive methodologies to generate listener reactions~\cite{ng2022learning,ng2023can,zhou2022responsive}.
These autoregressive methods have inherent structural drawbacks. In particular, during the inference stage, autoregressive models are sensitive to accumulating prediction errors. 
These accumulated errors can lead the generated sequence to significantly differ from the desired target and produce incoherent or unnatural listener responses.
Although a NAR(non-autoregressive) generation approach~\cite{song2023emotional} has been explored, it comes with its own constraints. The reported results are limited to short durations, and extending the response length requires a larger codebook size, which may limit the scalability of the approach.

To address the limitations of existing listener head generation methods, we propose \emph{DiffListener} that generates longer listener responses, which is challenging but more practical, in a non-autoregressive manner and fixed codebook size. 
First, we train a VQ-VAE\cite{van2017neural} model to learn a discrete codebook that encodes listener-specific response patterns. Second, we employ a discrete diffusion model~\cite{gu2022vector} to generate diverse and varied listener responses while preserving the codebook representation. 
The model performs the denoising diffusion process on the codebook tokens, which represent the listener's responsive reactions.
Prior research~\cite {ng2022learning,zhou2022responsive,song2023emotional} has primarily focused on the speaker's facial and audio cues to generate listener responses. 
However, this approach may overlook crucial lexical context. 
Our method incorporates textual information to consider this aspect.
Also, compressing speaker modalities into a condition module might result in a loss of temporal rhythmic information.
To overcome it, we incorporated the speaker's facial differential information, which helps maintain temporal rhythmic information, potentially enhancing the naturalness and coherence of generated reactions.
In our experiments, DiffListener outperforms the existing baselines in both quantitative and qualitative evaluations. These results demonstrate the effectiveness of our proposed approach.

In summary, the key contributions are as follows:
\begin{description}
    \item[$\bullet$] We propose DiffListener which is a novel non-autoregressive framework for listener head generation.
    To the best of our knowledge, it is the first to apply the discrete diffusion to listener generation task.
    \item[$\bullet$]
    We propose utilizing the facial differential and text information
    into the non-autoregressive generation framework to provide more context information.
    \item[$\bullet$] DiffListener achieves state-of-the-art performance on the listener head generation task, generating longer sequences while preserving high quality and relevance. 
\end{description}

\section{Related Work}
\subsection{Listener Head Generation}
VICO~\cite{zhou2022responsive} 
proposes the LSTM-based model which generates the listener's responsive reactions based on the speaker's facial and audio information.
L2L~\cite{ng2022learning} points out the non-deterministic properties of the listener's reaction. They propose using a codebook to represent the listener's reaction.
ELP~\cite{song2023emotional} proposes using multiple codebooks based on the estimated emotion from the speaker.
LM-Listener~\cite{ng2023can} utilizes the large language model to generate the listener's reaction only using the text information of the speaker. 
Most of the existing listener head generation models~\cite{ng2022learning, ng2023can, zhou2022responsive} utilize the autoregressive approach. However, this approach has drawbacks such as accumulated error problems, so we propose the DiffListener to solve this limitation in non-autoregressive manner.

\subsection{Diffusion Model}
The denoising diffusion probabilistic model~\cite{sohl2015deep} has shown strong performance not only in the image generation~\cite{ho2020denoising,ho2022cascaded,epstein2023diffusion}
but also in the other various generation tasks~\cite{kong2020diffwave,gong2022diffuseq, ho2022imagen}. 
The Argmax Flow~\cite{hoogeboom2021argmax} extends the diffusion model to categorical random variables with transition matrices. 
The VQ-Diffusion~\cite{gu2022vector} improves the discrete diffusion and applies it to the image generation task, which has also been applied in various tasks \cite{yang2023diffsound,li2023generalized,han2024clip}. To the best of our knowledge, this is the first time discrete diffusion is applied to the listener generation task.
\begin{table*}[t]
\centering
\caption{Comparision of our model with baselines in Trevor and Stephen dataset. Bold represents the best. 
$\downarrow$ indicates lower is better, no arrow indicates that closer to GT is better. GT denotes the Ground Truth.}
\label{tab:overall metric}
\resizebox{0.9\textwidth}{!}{%
\begin{tabular}{lcccccccccc}
\hline
            & \multicolumn{5}{c}{Trevor}                    & \multicolumn{5}{c}{Stephen}                  \\ \hline
Models      & L2$\downarrow$    & FD$\downarrow$    & P-FD$\downarrow$  & Diversity & Variation & L2$\downarrow$   & FD$\downarrow$    & P-FD$\downarrow$  & Diversity & Variation \\ \hline
GT          &       &       &       & 2.59      & 0.11      &      &       &       & 3.81      & 0.18      \\ \hline
VICO~\cite{zhou2022responsive}        & 0.41 & 17.94 & 19.22 & 2.31      & 0.08      & 0.71 & 31.43 & 33.64 & 2.98      & 0.11      \\
L2L~\cite{ng2022learning} & 0.62           & 25.28          & 27.11          & 4.66 & 0.28 & \textbf{0.65} & 28.69          & 30.33          & 2.79 & 0.09 \\
LM-Listener~\cite{ng2023can} & 0.69  & 28.67 & 30.56 & 4.77      & 0.29      & 0.79 & 37.57 & 39.21 & 2.34      & 0.09      \\
Ours               & \textbf{0.40} & \textbf{15.75} & \textbf{17.22} & 2.96 & 0.10 & \textbf{0.65} & \textbf{26.47} & \textbf{28.47} & 3.56 & 0.14 \\ \hline
\end{tabular}%
}
\end{table*}

\section{Methodology}
\subsection{Problem Definition}
To represent the precise 3D facial expressions and motions from video frames, we utilize the 3D Morphable Face Model(3DMM)~\cite{blanz2023morphable, kittler20163d} for each frame. Through this process, we can get the coefficient that corresponds to the facial expression $\beta_t \in \mathbb{R}^{d_m}$ where $d_m$ is the dimension of expression coefficient, head rotation $R_t \in SO(3)$, and identity-specific shape\cite{zollhofer2018state}. We discard the shape coefficient to model the representation that is independent of individual identity\cite{ng2022learning}. 
Our facial information at time $t$ can be represented by the concatenation of the facial expression and rotation coefficients $f_t = [\beta_t, R_t]$.
Let $F^{S}=\{ f^{S}_1, f^{S}_2, \cdots,f^{S}_{T-1},f^{S}_T \}$ represent the sequence of the speaker's facial information across T frames, and let $A^{S}$, $F_\Delta^{S}$, and $W^{S}$ represent their corresponding audio sequence, differential facial information, and text information, respectively. 
The differential facial information of the speaker is calculated taking the difference between the facial information in each time step and the previous time step:
$F_\Delta^{S} = \{f^S_x-f^S_{x-1}\mid 1\leq x\leq T\}$.
When DiffListener is given the speaker's information, the listener's responsive head sequence of the corresponding length $F^{L}$ can be generated:
\begin{equation}
    F^{L} = DiffListener(F^{S},A^{S},F_\Delta^{S},W^{S})
\end{equation}

\begin{figure}[t]
\centering
\includegraphics[width=1\columnwidth]{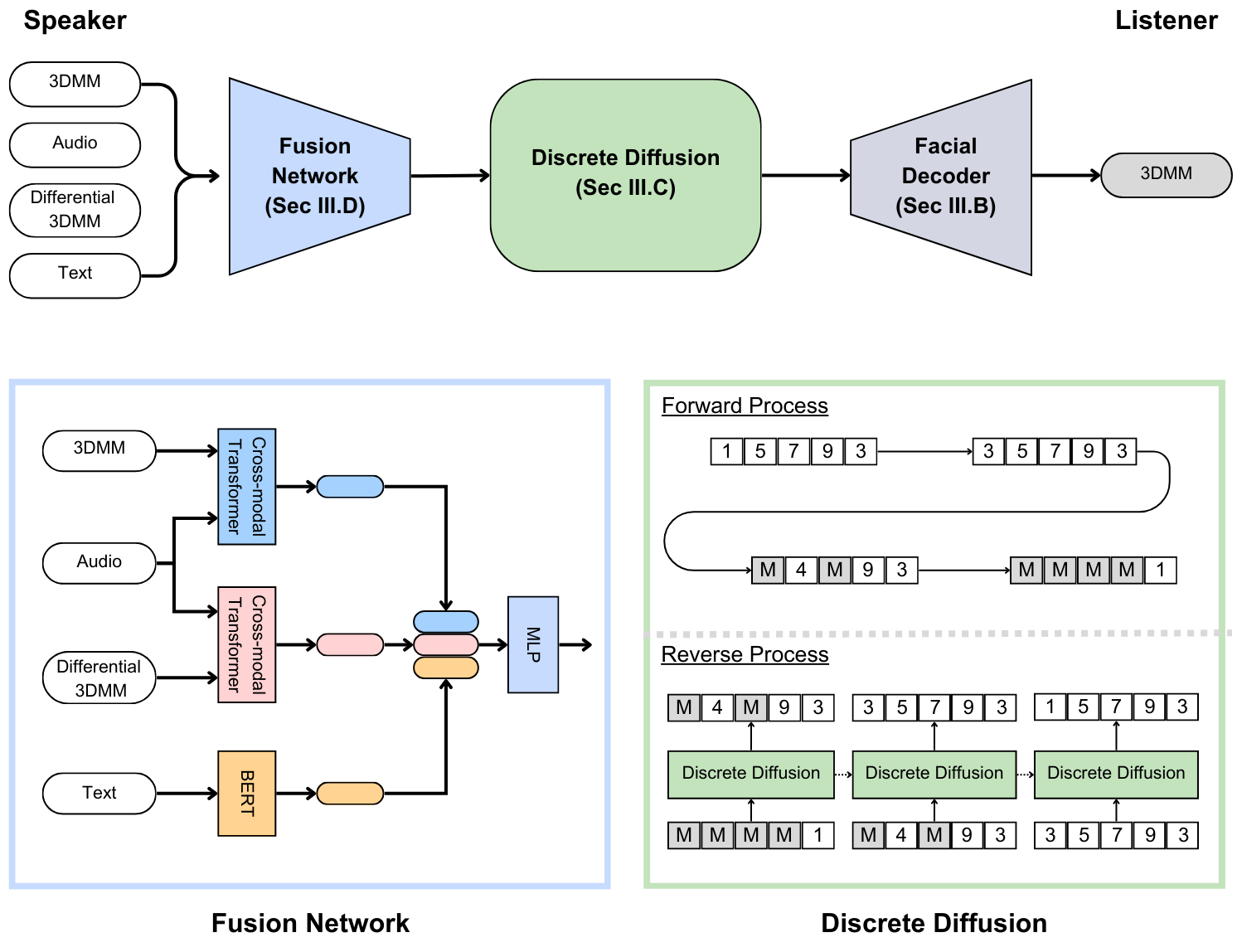}
    \caption{\textbf{The overview of DiffListener.} 
    The discrete diffusion model takes the fused representation and generates the listener's response sequence tokens. These tokens are then passed through the VQ-VAE decoder to obtain the listener's 3DMM coefficients.
    }
    \label{fig:Architecture overview}
\end{figure}
\subsection{Quantizing Listener Motion}
We use the VQ-VAE~\cite{van2017neural} to quantize the listener's facial information.
The VQ-VAE consists of facial encoder $\mathit{E}$, facial decoder $\mathit{D}$ and codebook $C = \{c_k\}^K_{k=1} \in \mathbb{R}^{K \times d_z}$ where $K$ is the size of the codebook and $d_z$ is the dimension of each code. Given the sequence of listener facial information $F^{L}=\{ f^{L}_1, f^{L}_2, \cdots, f^{L}_{T-1}, f^{L}_T \} $ across T frames,
we encode $F^{L}$ into specific representation through the facial encoder ${Z}=\mathit{E}(F^{L}) \in \mathbb{R}^{\frac{T}{\tau} \times d}$  where $\tau$ represents the ratio of downsampling and $d$ represents the dimension of the representation.
After that the vector quantizer $\mathit Q$ maps each ${Z}$ elements to its closest codebook entry.
Finally, the facial decoder reconstructs the sequence of the listener's facial information. We utilize the combination of losses to train the VQ-VAE:
\begin{equation}
    L_{embed} = \sum_{t=1}^{T/\tau}\parallel z_t-sg[c_i] \parallel^2
\end{equation}
\begin{equation}
    L_{reconstructed} = \sum_{t=1}^T \mathit L_{1}^{smooth} (\hat f_t - f_t)
\end{equation}
\begin{equation}
    L_{velocity} = \sum_{t=1}^{T-1} \mathit L_{1}^{smooth} (\hat f_{t+1} - \hat f_t, f_{t+1} - f_t)
\end{equation}
where $sg$ is the stop-gradient operation, and $\mathit L_{1}^{smooth}$ is the L1 smooth loss function. The total loss is a weighted sum of these losses. 

\subsection{Discrete Diffsuion Model}
We use VQ-Diffusion\cite{gu2022vector} to generate the listener response conditioned on the speaker's representation.
To get the speaker's representation, we utilized the speaker's information $I^S$ which consists of $F^{S}, A^{S}, F_\Delta^{S}, W^{S}$.
This information is then processed through the fusion network described in Section~\ref{sec:fusion_net} to generate the corresponding speaker's representation $R^S$. 
Let $\mathit{x}$ be a token sequence of length $T/\tau$ representing the listener's response, where each token is from the codebook.
During the forward diffusion process, the data $\mathit{x}$ is progressively corrupted through a fixed Markov chain $q(x_t|x_{t-1})$, which randomly replaces some tokens of $\mathit{x_{t-1}}$ over the diffusion time steps $T_d$. The reverse diffusion process then restores the data from $\mathit{x_{T_d}}$ based on the architecture.
The VQ-Diffusion\cite{gu2022vector} operates based on a transition probability matrix. 
The transition probability matrix $[Q_t]_{mn} = q(x_t=m \mid x_{t-1} = n) \in R^{K \times K}$ that represents the probabilities of $x_{t-1}$ transit to $x_t$.
To get better reverse estimation, a [Mask] token is introduced, expanding the transition matrix to $Q_t \in \mathbb{R}^{(K+1) \times (K+1)}$~\cite{gu2022vector}. 
To estimate the posterior transition distribution $q(x_{t-1} \mid x_t,x_0)$, we train the denoising network $p_\theta(x_{t-1}\mid x_t,R^s)$.
The network is trained to minimize the variational lower bound~\cite{sohl2015deep}:
\begin{align}
    L_{vlb} = \sum_{t=1}^{T_d-1}[D_{KL}[q(x_{t-1} \mid x_t,x_0) \parallel p_\theta(x_{t-1} \mid x_t, R^s)]] \notag \\
    +D_{KL}(q(x_{T_d}\mid x_0) \parallel p(x_{T_d}))
\end{align}
where $p(x_{T_d})$ refers to the prior distribution of timestep $T_d$. To get better results we utilize the reparameterization trick~\cite{gu2022vector}. 
This leads to a more noiseless distribution at each step. 
\begin{equation}
    L_{x_0} = -logp_\theta(x_0\mid x_t,R^s)
\end{equation}
where $L_{x_0}$ refers the auxiliary denoising loss, which can be further improved when combined with the $L_{vlb}$.

\subsection{Fusion Network}\label{sec:fusion_net}
Initially, we extract the text features by using the pre-trained language model BERT~\cite{devlin2018bert} and audio features using Mel-frequency cepstral coefficients (MFCC). 
The fusion network consists of two components, as illustrated in Figure~\ref{fig:Architecture overview}. 
The first module focuses on fusing audio and 3DMM features. Cross-modal attention is computed using the queries $Q_A$ from $A^S$, and the keys $K_F$ and values $V_F$ from $F^S$. The second module handles the fusion of audio with differential 3DMM features. In this module, we set the queries $Q_{F_\Delta}$ from $F_\Delta^S$ and the keys $K_A$ and values $V_A$ from $A^S$. 
Finally, we concatenate each of the fused representations along with the text representation and feed the combined representation into an MLP layer for further processing.
\section{Experiment}
\begin{table}[t]
\centering
\caption{Ablation study results based on codebook size at Trevor dataset. GT denotes the Ground Truth.}
\label{tab: ablation codebook size}
\resizebox{0.9\columnwidth}{!}{%
\begin{tabular}{cccccc}
\hline
         Codebook size  & L2$\downarrow$   & FD$\downarrow$    & P-FD$\downarrow$  & Diversity & Variation \\ \hline
GT & & & &2.59 &0.11 \\ \hline 
\textit{128}       &  0.47    &  17.61     & 19.28      & 3.63      & 0.17      \\
\textit{256}   & \textbf{0.40}  & \textbf{15.75} & \textbf{17.22} & 2.96      & 0.10    \\
\textit{512} & 0.48 & 18.28 & 20.01 & 3.88      & 0.18    \\
\hline
\end{tabular}%
}
\vspace{-0.2cm}
\end{table}

\subsection{Dataset}
Our research objective is to generate longer identity-specific listener responses. To achieve this goal, we select the dataset~\cite{ng2023can,ng2022learning}, which provides sufficiently long sequences of listener behaviors while maintaining individual identity characteristics.
We preprocess the dataset by setting the listener’s response period to 8 seconds (240 frames), assuming that this duration may be sufficient to understand contextual information through text data~\cite{ng2023can}.
We clip each video to 8 seconds and employ a sliding-window approach to generate more data.
Based on this preprocessing, we found sufficient data from Trevor and Stephen identities.
To extract text from audio data, we utilize the Whisper~\cite{radford2022robust}.

\subsection{Experimental Setup}
Following the previous work \cite{ng2023can,ng2022learning}, 
we use the following evaluation metrics for quantitative evaluation. L2, FD(Frechet Distance), P-FD (Paired FD) which is the distribution distance between listener and speaker, Diversity, and Variation.
We use the L2L~\cite{ng2022learning}, VICO~\cite{zhou2022responsive}, and LM-Listener~\cite{ng2023can} as our baselines.
The ELP~\cite{song2023emotional}, which is based on the non-autoregressive approach, does not release the code, so we are not able to compare ours with it.
We use $K=256$, $T=240$, $T_d=100$, $d_z=512$, $\tau=8$ with a batch size = $256$ when training on the Trevor dataset~\cite{ng2023can}, and batch size 64 when training on the Stephen dataset~\cite{ng2023can}.
The value $K=256$ is chosen by its superior performance, as shown in Table~\ref{tab: ablation codebook size}.
During DiffListener training, we apply early stopping with a patience of 5 epochs. For VQ-VAE training, we set the weights of each loss term as follows: $0.02$ for $L_{embed}$, $1$ for $L_{reconstructed}$, and $0.05$ for $L_{velocity}$.

\subsection{Quantitative Comparison}
Table~\ref{tab:overall metric} presents the overall performance of our model and the baselines. Our model achieves the lowest L2, FD, and P-FD scores for both Trevor and Stephen's datasets, indicating the highest realism in generating listener motions and the greatest synchrony with the speaker.
Our model achieves superior performance in terms of diversity and variation scores in most cases. 
For the Trevor dataset, the diversity metric is slightly less close to the ground truth compared to the VICO model. 
However, the difference is minimal, and our model achieves a higher diversity score and lower L2, FD, and P-FD scores than VICO. This indicates that our model can generate more realistic and diverse results.

\begin{table}[t]
\centering
\caption{
Ablation study results based on speaker modalities at trevor dataset.
}
\label{tab:ablation modality}
\resizebox{0.95\columnwidth}{!}{%
\begin{tabular}{lccccc}
\hline
         & L2$\downarrow$   & FD$\downarrow$    & P-FD$\downarrow$  & Diversity & Variation \\ \hline
GT       &      &       &       & 2.59      & 0.1127      \\ \hline 
w/o Diff \& Text   & 0.50  & 20.77 & 22.27 & 3.18      & 0.1223    \\
w/o Diff & 0.44 & 17.17 & 18.71 & 3.21      & 0.1247    \\
w/o Text & 0.41   & 16.17   & 17.66    & 3.05    & 0.1095 \\
Ours & \textbf{0.40} & \textbf{15.75} & \textbf{17.22} & 2.96 & 0.1027 \\ \hline
\end{tabular}%
}
\end{table}

\begin{figure}[t]
    \centering    \includegraphics[width=\columnwidth]{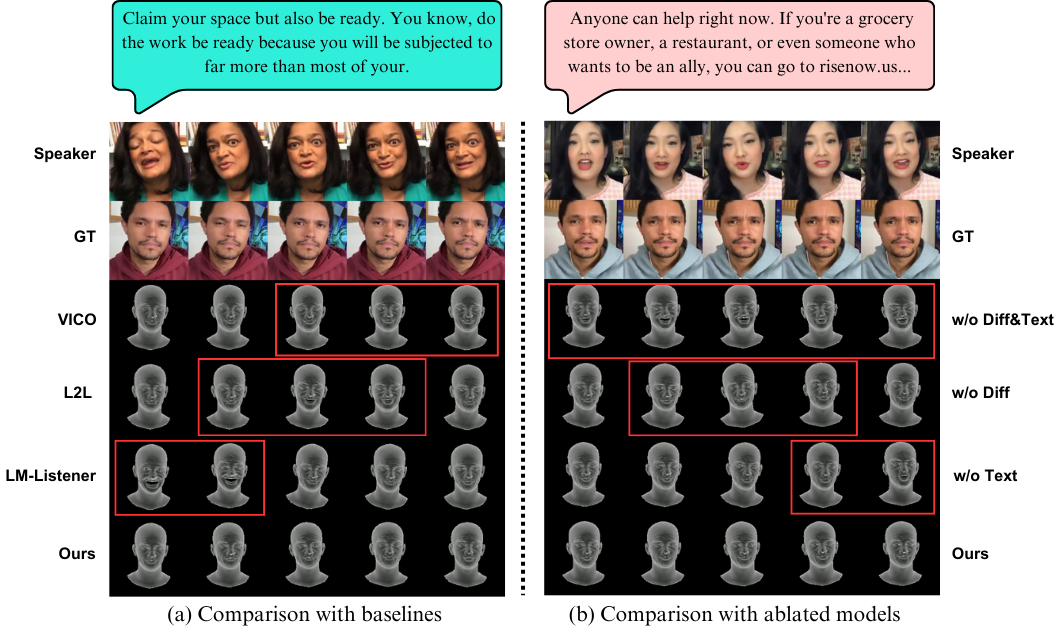}
    \caption{The visualization comparison with baselines and ablated models at Trevor dataset. The blue and pink boxes include the speaker's utterance.}
    \label{fig:  visualize trevor result}
    \vspace{-0.5cm}
\end{figure}

\begin{table}[t]
\centering
\caption{User study results based on the comparison of our model with baselines.}
\label{tab:userstudy}
\resizebox{0.95\columnwidth}{!}{%
\begin{tabular}{lcccccc}
\hline
                    & \multicolumn{3}{c}{Trevor} & \multicolumn{3}{c}{Stephen}   \\ \hline
                    & Win     & Tie     & Lose   & Win     & Tie     & Lose      \\ \hline
Ours vs. Vico~\cite{zhou2022responsive}        & \textbf{51\%}    & 23\%    & 26\%   & \textbf{65\%}    & 11\%    & 24\%      \\
Ours vs. L2L~\cite{ng2022learning}         & \textbf{50\%}    & 11\%    & 39\%   & \textbf{41\%}    & 25\%    & 34\%      \\
Ours vs. LM-Listener~\cite{ng2023can} & \textbf{67\%}    & 16\%    & 16\%   & \textbf{47\%}    & 12\%    & 41\%      \\
 \hline
\end{tabular}%
}
\vspace{-0.5cm}
\end{table}

Table~\ref{tab:ablation modality} presents the results of our ablation study on different modality conditions.
By incorporating text into a model without differential and text information (w/o Diff \& Text),
we observed approximately $17.33\%$ decrease in FD score 
(w/o Diff).
Only by incorporating differential 3DMM information into the model to provide rhythmic temporal information, we observed an approximately $22.16\%$
decrease in the FD score (w/o Text).
Furthermore, integrating the differential 3DMM value and text, we observed a $24.1\%$ decrease in the FD score and $22.6\%$ decrease in the P-FD score (Ours).
This indicates that the combination of differential information with audio, 3DMM, and text data enhances the model's ability to understand the context information, thereby enabling it to generate more appropriate responses.

\subsection{Qualitative Comparision} 
In the Listener Generation task, it is important to generate results that closely match the ground truth.
Figure~\ref{fig: visualize trevor result}(a) presents a visual comparison with the baseline models. The baselines sometimes fail to generate appropriate responses, as indicated by the red boxes. While the ground truth sample is in a neutral state, other baselines sometimes show a smile. In contrast, our model demonstrates more appropriate responses in both head pose and facial expression.
This result may come from the advantages of DiffListener.
First, NAR approach can avoid the problem of accumulated errors. It makes our model's results more robust during the inference.
Second, using various modalities provides sufficient contextual information.
Figure~\ref{fig: visualize trevor result}(b) presents a visual comparison with ablated models. When sufficient contextual information is provided, the model generates more appropriate listener responses. However, when some modalities are excluded, it produces inappropriate responses, as indicated by the red boxes.
In addition, we conduct a user study to evaluate human preferences on Amazon Mechanical Turk. We randomly select 25 videos from each of the identity datasets (total 50 videos) and visualize each result as a grayscale mesh video using EMOCA~\cite{Danecek_2022_CVPR} because mesh videos provide a more intuitive way to evaluate facial and head movements compared to photorealistic videos~\cite{song2023emotional}. If photorealistic video is required, it can be generated using a renderer model~\cite{ren2021pirenderer}.
Given the (speaker, ours, baseline) tuple of samples, 20 people were asked to choose the video that appeared to be listening and paying more attention to the speaker. The results are presented in Table~\ref{tab:userstudy}. Our model is more preferred than the baselines in both datasets. More samples can be found on our demo page\footnote{\url{https://siyeoljung.github.io/DiffListener/}}.
\section{Conclusion}
In this work, we propose DiffListener, a novel approach that generates realistic and diverse listener responses in a non-autoregressive manner using a discrete diffusion model.
Unlike previous work, our method can generate longer responses while maintaining a fixed codebook size. 
To better synchronize with the speaker, we introduce a novel speaker modality (the speaker's facial motion differential).
Through experiments, we demonstrated that our approach outperforms existing baselines in terms of realism, diversity, and synchrony with the speaker's motions.
This work represents a significant step forward in achieving more natural and context-aware listener generation.

\bibliographystyle{ieeetr}
\bibliography{reference}

\end{document}